\documentclass{IEEEtran} 
   
\usepackage{url}
\usepackage[hidelinks]{hyperref}
\usepackage[utf8]{inputenc}
\usepackage{graphicx}
\usepackage{amsmath}
\usepackage{booktabs}
\usepackage{algorithm}
\usepackage{algpseudocode}

\usepackage{listings}
\usepackage{color}

\begin{document}
%
\title{ELO System for Skat and Other Games of Chance} 
\author{Stefan Edelkamp, AI Center at CTU Prague, Czech Republic
}
\maketitle
\begin{abstract}
Assessing the skill level of players to predict the outcome and to rank the players in a longer series of games is of critical importance for tournament play. 
Besides weaknesses, like an observed continuous inflation, through a steadily increasing playing body, the ELO ranking system, named after its creator Arpad Elo, has proven to be a reliable method for calculating the relative skill levels of players in zero-sum games. 
 
The evaluation of player strength in trick-taking card games like Skat or Bridge, however, is not obvious. 
Firstly, these are incomplete information 
partially observable games with more than one player,
where opponent strength should influence the scoring as it does in 
existing ELO systems. Secondly, they are game of both skill and chance, so that besides the playing strength the outcome of a game also depends on the deal. Last but not least, there are internationally established scoring systems, 
in which the players are used to be evaluated, 
and to which ELO should align. 
Based on a tournament scoring system, we propose a new ELO system for Skat to overcome these weaknesses.
\end{abstract}

\lstset{language=C++,morekeywords={array,constraint,var,forall,sum,solve,minmize,decreasing,domain},basicstyle=\ttfamily,keywordstyle=\color{blue}\ttfamily,
literate=%
  {+}{{{\color{red}+}}}1
  {!}{{{\color{red}!}}}1
  {*}{{{\color{red}*}}}1
  {/}{{{\color{red}/}}}1
  {=}{{{\color{red}=}}}1
  {|}{{{\color{red}|}}}1
  {\%}{{{\color{red}$\%$}}}1
  {<}{{{\color{red}<}}}1
  {~}{{{\color{red}$\sim$}}}1
  {\&}{{{\color{red}\&}}}1 
  }

\newtheorem{definition}{Definition}
\newtheorem{theorem}{Theorem}
  \setlength{\tabcolsep}{5pt}

\noindent

\section{Introduction}
 
The ELO rating system~\cite{Elo} is an established method for calculating the relative skill levels of players in zero-sum games such as Chess. The playing strength of a Chess player, program or engine, reflects the ability to win against other players, given by a number. ELO is also used in Video Games, American Football, Basketball, Major League Baseball, Table Tennis, Scrabble and Diplomacy, and many other games. 

The ELO rating has been devised for games of skill,
usually two-player zero-sum deterministic games with full observability. The difference in the ratings between two players serves as a predictor of the outcome of a single game. Two players with equal ratings, who play against each other are expected to score an equal number of wins. A player whose rating is 100 points greater than their opponent's is expected to score 64\%; if the difference is 200 points, then the expected score for the stronger player is 76\%.

The players' ELO rating is represented by a number which changes depending on the outcome of rated games played. After every game, the winning player takes points from the losing one. The difference between the ratings of the winner and loser determines the total number of points gained or lost after a game. If the high-rated player wins, then only a few rating points will be deducted from the low-rated player. However, if the lower-rated player scores an upset win, many rating points will be added to his rating. The lower-rated player will also gain a few points from the higher rated player in the event of a draw, so that this rating system is self-correcting. Players whose ratings are too low (or too high) should, in the long run, get better (or worse) than the rating system predicts and, thus, gain (or lose) rating points until the ratings reflect their true playing strength.

The ELO ranking system is highly influential for tournament play. 
In Chess, for example, a world-wide live ranking list 
decides on tournament invitations, up selecting candidates for challenging the World Champion. Moreover, evaluating the performance level is useful for tracking progress of players and AIs in online game playing, and to compare the level of play with human strength. 

For games that include chance, like card games, however,
devising an appropriate ELO ranking system is much more challenging~\cite{Bewersdorff}. Due to the randomness in the deal, it takes more games to derive statistical meaningful results.
Such a novel ranking system has a huge impact, as it can be used for a evaluating tournaments and set-up chance-reduced, skill-emphasizing, much fairer ranking list of the players, to be awarded for their skill of play. It is very much needed for online platforms, as motivating aspects for the players it helps to producing high-score tables.  

In many countries, there are monetary and legal consequences, whether a game is considered to be 
based on skill or chance. For the usual setting of 
including both, extracting the skill factor of a game of chance is wanted. 
Existing ranking systems like average scoring values
are sensitive to the strengths of the opponent players, which cannot be compared easily to a global ranking.
Even when working towards this ultimate goal, by looking at the ELO ranking, however, we cannot 
expect an exact derivation of a probability of winning 
as done in the original ELO system. Instead, we
align the ELO ranking to an accepted scoring systems.

In this paper, therefore, we propose and implement a 
refined ELO system that covers most important aspects of such games, compensating for the strength of the opponents and the factor of chance. We use it for evaluating players performance in the experiments,
and discuss different design choices and parameterizations. The results are compared to AIs
self-play. We exemplify our considerations in for the game of Skat~\cite{Kupferschmid1,Kupferschmid2,Keller1,Keller2}, but the system can be adapted and applied to other card games and beyond.

We start with a motivation, the definition of known tournament score and introduce some approaches judgement of card value. For the new proposal of an ELO rating system we look into related work. The proposal is 
presented step, by step, starting from the existing
ELO system for two players, and its possible extension
to many players, we derive the general formula for
the inclusion of opponent strength and the alignment
to the existing scoring values, including the
known $K$ factor influencing the spread of values. 
Two different approaches to include chance are discussed changing the game value of a single game or, subsequently, in a series. The experiments evaluate a moderately large body of human played games, which are also replayed with
our considerably advanced Skat bots.

\begin{figure}
    \centering
    \includegraphics[width=6cm]{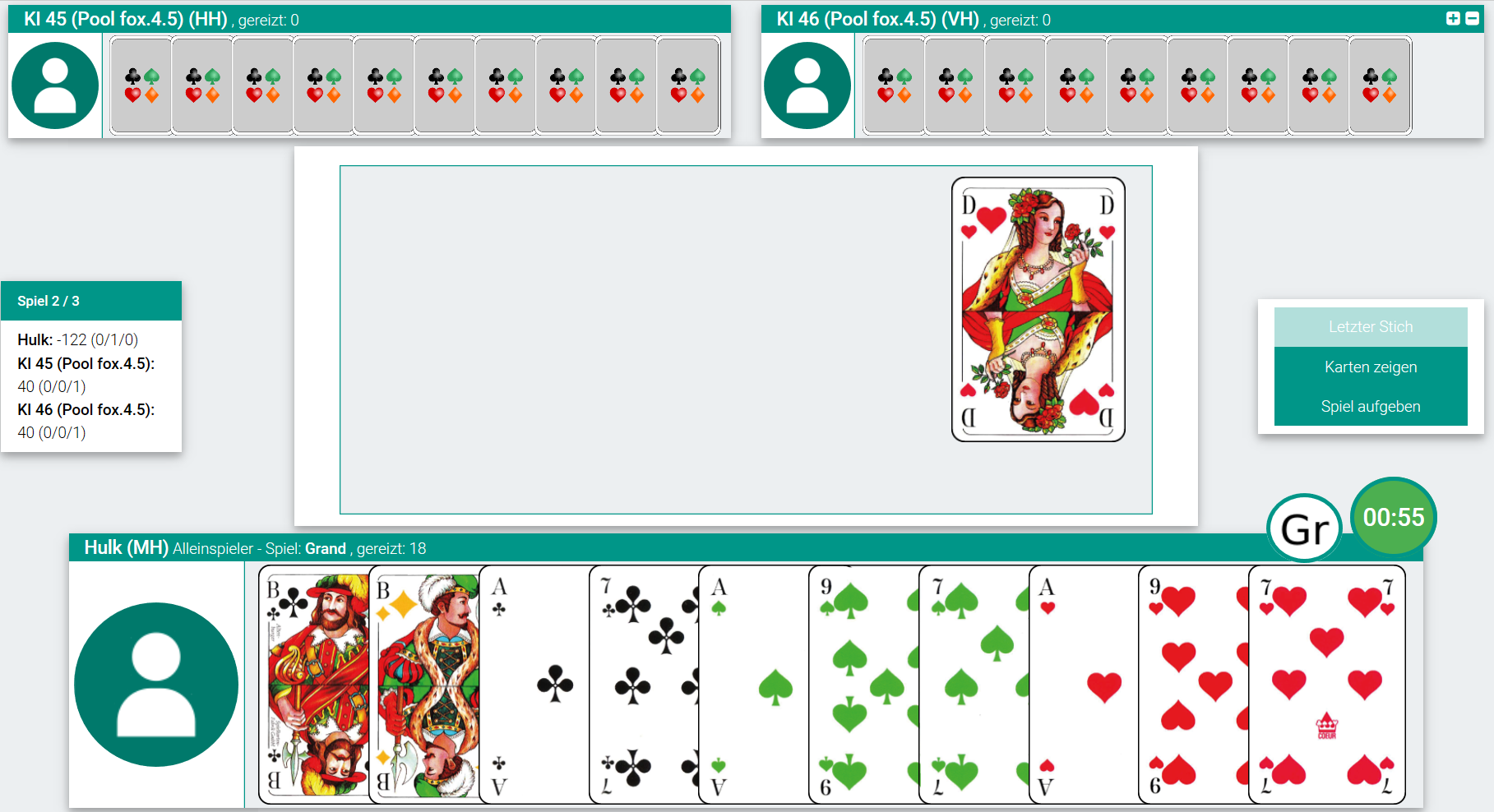}
    \caption{A game of Skat at start of trick-taking stage.}
    \label{fig:gameofskat}
\end{figure}

\section{Motivation}

Skat (Fig.~\ref{fig:gameofskat}) is one example for a game of mixed chance and skill~\cite{Quambusch,Grandmontagne,Harmel,Lasker1,Lasker2,Rainer}. 
There is a wide consensus 
that while in a single game the variation due to randomness in the deal is large, good players show their strength
eventually.
Skat is played with a deck of 32 cards of
which each player gets 10 in a given deal.
The left-over \emph{skat} cards are bid for.
The winner of
the bidding becomes the declarer, while
the other two are playing together as opponents. 
There are different types of contracts that can be declared and the two cards that after taking the
skat are put count to the declarers' score.
 
Skat is played in club, national and international 
tournaments, either online or \emph{over-the-table}, where the winning is determined either by number of wins, accumulated 
scoring or via the (extended) Seeger scoring system. Tournament results and rating lists often show average scoring results.
By the continuous growth in playing Skat, a fair, transparent, and objective ranking for the game is a demanding
necessity, however, due to the situation of being a 
three-person game with unknown card distributions, this is much more complicated than in Chess with its defined starting position.

So far, this has only been done based on the average points in the accepted scoring system, and, in the event of a tie, based on the larger number of games won.

Even if played for hobby and leisure, what is generally missing is a players' reference value, to match a player to an equally strong table. There are early approaches implemented in some Skat forums, where they take all players and then rank only the players who are advanced according to an internal analysis. There is also the need for a high level of confidence, especially if there are significantly fewer data sets behind. If there are only a few games play a good score can be a matter of simply having good cards.

One has tried to create a reference value using an average scoring, e.g., \emph{25 points / game} for a player, and of course adding some bonus at a higher player level, if you still win against strong opponents as the declarer.

How much is that, is the mathematical question, that we will be try to resolve.
What has generally been neglected so far is to include a measure for hand strength of a game in the evaluation. Because a game with $100\%$ winning probability almost everyone can win; but a game only with $50\%$ only the fewest. There are sufficiently tested, simple and complex  winning forecasts for games so that one could receive more points for winning a simpler game than for a difficult one.

To allow weaker players a better scoring,  
some online platforms try to dampen the 
scoring of good players by modifying the deal,
given worse cards to the stronger players. 
We strongly believe this is the wrong way of dealing with the problem, hard to make fair and 
transparent, and players experience this 
as being closer to a fraud.
It is better to include the handicap 
into the ranking function, so that weaker players
are awarded if they win against stronger opponents.

What we want is a ELO scoring system with a similar
outcome as the Seeger-sytem of DSKV/ISPA, which
compensates for the factor of chance, and normalizes wrt.\  the strength of the
opponents. This sould lead to a wide acceptance at least
for Skat enthusiasts, tournament directors, and last
but not least the Skat players. 
In the ELO value we want 

\begin{itemize}
    \item 
to include the factor of chance as a surplus or deduction wrt. 
estimated chance of winning.
\item 
to cope for 
the value describing the \emph{game size}, i.e., 
the mean value against a averaged comparison
value, computed once for a series. 
\item
an ELO list adapted accordingly; for a
starting value, every series of 36 games
3 new ELO numbers are computed.
\item 
if the player plays against strong opponents s/he
gets rewarded for the risk taken, 
while if s/he plays against weaker ones, s/he receives less bonus.
\item
to choose parameters to reach 
an asymptotic 1100 extended Seeger Pts/series for a good 
Skat player, and about 650 Pts/series 
for a bad player
\item a quick evaluation of thousands of games
to compared player performance in ELO rating list.
\end{itemize}

\subsection{Game of Skill or Game of Chance?}

We are aiming at a fair evaluation of the playing strength with reduced components of chance. 
Duersch, Lambrecht, and Oechsler~\cite{Duersch}
studied how much different sorts of  games are affected by chance. They took playing data (from \emph{Game-Duell} and different other sources), and compared
the statistics on ELO evolution applying an extension of the Chess formulas. They were not considering the actual input, i.e., for card game the deal.
Instead they provide a benchmark to determine, whether or not a game depends predominantly on chance, through a comparison to a game by randomly replacing 50\% of outcomes in Chess with coin flips. 
In their study by the evolution of ELO values they claimed that 50\% Chess still shows more skill elements than Skat, which on this measure for skill vs chance, is on par with
Backgammon and above Poker.
Even without knowing the game of Skat they found that it is quite
sensitive to the factors of card luck (which is undoubted 
by many players). In Skat, only
a considerably small database of
played games was taken. 

Borm and van der Genugten~\cite{Borm1,Borm2} propose measures that compare the performances of different types of players. In order to calculate which part of the difference in performance may be attributed to 
skill and which to chance, they include as a benchmark an informed hypothetical player who knows exactly which cards will be drawn.
All these approaches are not helpful to determine which player 
is pushed back or forth by fortune of having bad or good setting. 
And how to correct the ELO-computation accordingly.  
 
\section{Scoring Systems
}

The extended Seeger (aka Seeger-Fabian) system~\cite{Rainer} is an established scoring system for Skat to evaluate a series of $X$ games. Usually, we have $X = 36$. 
Folded games, for which there are no bids, are neglected. 


The evaluation score is based on the number of wins and losses of each player in the series, and the game value oft the games being played. For a single game $g$ of a player $P$, the outcome is $V(P,g)$, if the game is won for the declarer, and $-2 \cdot V(P,g)$, if it is lost. In a series of games $G = g_1,\ldots,g_k$ these values are added for each player, so that 
$V(A,G) = V(A,g_1) + \ldots + V(A,g_k)$.

For example, we have players $A$, $B$, and $C$ with 
profiles \\
\qquad $(\#wins(A,G), \#losses(A,G),V(A,G))$, \\
\qquad $(\#wins(B,G), \#losses(B,G),V(B,G))$, and \\
\quad $(\#wins(C,G), \#losses(C,G),V(C,G))$, respectively.

In the extended Seeger scoring System the evaluation strength $ES$ is defined as
$ES(A,G) = V(A,G) + 50 \cdot (\#wins(A,G) - \#losses(A,G)) + 40 \cdot (\#losses(B,G) + \#losses(C,G))$.
Similarly, for $B$ and $C$ we have 
$ES(B,G) = V(B,G) + 50 \cdot (\#wins(B,G) - \#losses(B,G)) + 40 \cdot (\#losses(A,G) + \#losses(C,G))$
and 
$ES(A,G) = V(A,G) + 50 \cdot (\#wins(C,G) - \#losses(C,G)) + 40 \cdot (\#losses(A,G) + \#losses(C,G)).$ 

The outcome of a series of $36$ games might be as follows
$V(A,G) = 273$, $\#wins(A,G) = 8$, $\#losses(A,G) = 1$, 
$V(B,G) = 152$, $\#wins(A,G) = 12$, $\#losses(A,G) = 4$, 
$V(C,G) = 495$, $\#wins(A,G) = 11$, $\#losses(A,G) = 0$.
Then, we have $ES(A,G) = 273 + 350 + 160 = 783$, $ES(B,G) = 152 + 400 + 40 = 592$, and $ES(C,G) = 495 + 550 + 200 = 1295$.
In a set of equally strong players, value $\ge 1000$ is considered to be the result of strong play. However, it could also be the result of very good cards.


\section{Hand Strength
}

Determining luck in the cards resorts
to knowing and comparing with the strength of the hand
or, even better, to the estimated winning probability.
None of the values are exact, but help to 
predict the outcome of the game.

Walter von Stegen proposed to evaluate the strength of the given hand using the points system an expert Skat player, youtuber and teacher. It only applies to trump games. Roughly speaking, it counts the number of Js with 2 points, and the number of trump and high-cards (As or 10s) with 1 point (high trump cards are counted twice). Additionally,
strong Js and low bids are awarded. 



Thomas Kinback
is a professional Skat player, teacher,
and author~\cite{Kinback}. 
He developed another counting system for hand strength, 
which is explained in greater depth by~\cite{Rainer}.
The hand strength value differs in grand and suit games, 
and includes a measure for tricks going home. Moreover, different 
configurations of cards in a suit as well as being in forehand get a surplus. 


We used an accurate prediction of the declarer's probability 
to win. 
More precisely, we devise 
a winning probability $\mbox{\em Prob}(h,s,t)$ including the Skats $s$ together with hand $h$. 
These probabilities are used in the first stages of the Skat game: bidding, skat taking and game selection, and skat putting. 
The expert games can easily be analyzed statistically,
which we do for all of the kinds of games being played.
There is, however, a fundamental difference in estimating the winning probability in null, and trump games.
For null games, we estimate the winning probability $\mbox{\em Prob}(h,c)$ in each suit $c$ separately. 
The winning probability, then, is  
$\mbox{\em Prob}(h) = \mbox{\em Prob}(h,clubs) \cdot \mbox{\em Prob}(h,spades) \cdot \mbox{\em Prob}(h,hearts) \cdot \mbox{\em Prob}(h,diamonds)$.
For trump games, we consider the so-called winning parameters:
Free suits: the amount of non-trump suits that the player does not have in hand;
\emph{skat value}: the number of eyes in the skat, condensed into four groups; \emph{bidding value}: the three different final bidding values in the bidding competition, summarized into four values;
\emph{declarer‘s position}: the position of the solist in the first trick;
emph{trump length}: the number of trump cards in hand;
\emph{non-trump aces and ten}s: the number of non-trump high cards in hand;
\emph{Jack group}: set of jacks condensed into groups;
\emph{lost cards}: number of tricks expected to be lost;

Through statistical tests we were able to validate that these parameters have a significant influence on the profit or loss of the expert games and can therefore be used as essential attributes to accurately assess the probability of winning a grand or suit game. More precisely, the grand table is based on 7 to 8 of these winning parameters and the color game table on 9 winning parameters.

\section{ELO System for Zero-Sum Games}

We now recapitulate the ELO ranking system that is known from games like Chess, and its extension to multiple players.

\subsection{Two-Player Games}

The ELO ranking system for two player 
in a zero-sum game works as follows.
If Player $A$ has a rating of $R_A$  and Player $B$ has a rating of $R_B$, the formula for the expected score of Player A is 
$$E_A=1/ ( 1+10^{(R_B-R_A)/400)} )$$
Similarly, the expected score for Player $B$ is  
$$E_B=1/ ( 1+10^{(R_B-R_A)/400)} ).$$ 
Hence, for each $400$ rating points of advantage over the opponent, the expected score is $10 \times$ the opponent's one. 

When a player's actual tournament scores exceed their expected scores, the ELO system takes this as evidence that player's rating is too low, and needs to be adjusted upward. Similarly, when a player's actual tournament scores fall short of their expected scores, that player's rating is adjusted downward. 

Elo's original suggestion~\cite{Elo}, which is still widely used, was a simple linear adjustment proportional to the amount by which a player over-performed or under-performed their expected score. The maximum possible adjustment per game, called the $K$-factor, was set at $K = 16$ for masters and $K = 32$ for weaker players.
Supposing Player $A$ was expected to score $E_A$ points but actually scored $S_A$ points. Updating that player's rating is by setting $R_A \leftarrow R_A + K \cdot (S_A - E_A).$

\subsection{Multiple Player Games}


If we stick to the simple case where the outcome of a $k$-player matchup is one winner and $(k-1)$ losers, then it seems incorrect to simply update player 1's rating on the basis of $R_1$ vs. average $(R_2, R_3, \dots, R_k)$.

The reason this seems invalid is that it's clearly more difficult to beat player $X$ against whom you have a $60\%$ chance of victory, followed by player $Y$ against whom you also have a $60\%$ chance of victory, than it would be to defeat player $X$ only. The former shows a greater consistent performance.

Another method would update the rating $(k-1)$ times treating it as though you defeated all $(k-1)$ opponents individually one after the other. This doesn't seem to have the same problem as the first way of doing it, but suffers from the issue of deciding which order to compute the rating updates in. A tweak on this approach is to take the average of all the opponents ratings and then compute the update $(k-1)$ times on that average rating. This avoids the order issue but seems to fail due to the fact that winning a $3$-player game against a player you have a $1\%$ chance against and another player you have a $99\%$ chance against seems much more difficult than winning a match against two players against both of which you are expected to score $50\%$.

A third way of doing it would be increasing the total score of a game. An chess game basically is a battle over one point. a draw means both sides won half a point, a win means the winner took the full point and the loser took none. possibly with k players, there would be $k/2$ points up for grabs which would mean that for the $S_A$ you'd have values up to $k/2$ instead of just $1$ for a win. applying $S_A = k/2$ to the formula and using average (opponent's ratings) for the rating, is another possible method, though it still suffers from the issue involving the difference between defeating a super strong and super weak player vs defeating two even players.

When two players have ratings $R_A$ and $R_B$, determined via the ELO model for two players, we have computed their expected scores against each other:
$$E_A = 1 / [ 1 + 10^{(R_B - R_A)/400} ]$$
and symmetrically:
$$E_B = 1 / [ 1 + 10^{(R_A - R_B)/400} ]$$
Their ratings are updated by comparing the result of the game with what was expected, e.g.,
$$R_A \leftarrow R_A + K \cdot (S_A - E_A),$$
where $S_A$ is the score of the game (1 for a win, $0$ for a loss, $.5$ for a draw, etc.) and $K$ is a \emph{volatility factor} that is equal to the maximum number of rating points that can be gained/lost in a game.
Symmetrically, whatever was added to $R_A$ will be subtracted from $R_B$.

One can rephrase ELO the following way: The probability of a player winning the game is modeled as being proportional to $10^{rating/400}$. Of course, you need to normalize the probabilities by dividing by the sum of these numbers. This is the same formula as before:
$$E_A = 10^{R_A/400} / (10^{R_A/400} + 10^{R_B/400}).$$
Dividing numerator and denominator by $10^{R_A/400}$ we get
$$E_A = 1 / (1 + 10^{(R_B-R_A)/400}).$$

This rephrasing is a lot easier to expand to more players: Each one has a probability proportional to $10^{rating/400}$ of winning the game.
$$E_A = 10^{R_A/400} / (10^{R_A/400} + 10^{R_B/400} + 10^{R_C/400}).$$

The update rule does not need to change.

\section{Significance and Related Work}

Gaming is a multi-billion dollar industry~\cite{Croson}, and games of chance (in contrast to games of skill) are often prohibited, or at least tightly regulated in many jurisdictions over the globe. 
Thus, the question, whether a game predominantly depends on skill or chance, has important legal and regulatory implications~\cite{Cabot,DeDonno,Fiedler}. 
Öchsler, Dürsch and Lambrecht~\cite{Duersch} 
suggest a new empirical criterion for distinguishing games of skill from games of chance. All players are ranked according to a \emph{best-fit} ELO algorithm. The wider the distribution of player ratings are in a game, the more important is the role of skill. Most importantly, they provide a new benchmark (50\%-Chess) that allows to decide, whether games predominantly depend on chance, as this criterion is often used by courts. They have applied the method to large datasets of various games (e.g. Chess, Poker, Backgammon)
\cite{Dreef,Camerer}.

The findings indicate that most popular online games, including Poker, are below the threshold of 50\% skill and thus depend predominantly on chance. In fact, Poker contains about as much skill as chess when $75\%$ of the Chess results are replaced by a coin flip.

According to Marco Lambrecht~\cite{a1}, in their studies, Skat was below der threshold of 50 percent skill, thus depending more on chance. Wrt.\ Poker, with lesser skill requirements compared to Skat he writes: \emph{nevertheless, skill prevails in the long term. Our evaluations show: After about one hundred games, a Poker player who is one standard deviation better than his opponent would have a 75 percent probability of winning more games than his opponent.} 

We came across the topic of devising an ELO scoring system when playing on Michael Buro's Skat platform, which offers a ELO-type of ranking and when we asked him about certain peculiarities. He couldn't remember how that number had been defined exactly, but we got some impressions on its pros and cons.
One thing had to be given to Buro‘s ELO numbers: they expresses the table strength situation relatively well, depending on whether you were winning or losing against stronger or weaker opponents (in terms of their ELO number). 

The value of the cards, however, was not included in the computation, which is of course always a drawback with long streaks of bad luck or luck. Assessing the skill level based on the points scored in relation to the value of the cards turns out to be an absolutely important aspect, such a corrected analysis says a lot about the true and luck-adjusted performance of a player.
Therefore, we find it fundamentally interesting to consider an ELO number as discussed above as a supplementary, alternative rating number.
The methodology how this is calculated in games of no chance like Chess is well understood, but there are some clear differences anyway, so we made up own thoughts and could propose the following algorithm.

\section{ELO System for Games of Chance}

To extend the ELO System to non-zero sum games and games of chance is advanced.
We kick-off by the usual scoring system used in tournaments, the extended
Seeger system, with a ranking along the average score.

\begin{table}[t]
    \scriptsize
    \begin{tabular}{c|c|cc}
Player  & Number of & Aver. Score & Average Score \\  
(anonymous) & Games & (per Game) & (per 36 series) \\ \hline
e15352892b7a6cd14e7dedaff97672ec & 195 & 30.75 & 1107.15 \\
8afc83bd59fa1ed31b88d6d4f12286be & 89 & 30.11 & 1084.31 \\
18b056ef4c0a2b11464af83bcf69b98d & 207 & 29.90 & 1076.58 \\
949df8dc794077d58b3628bed8f2d8b0 & 10  & 29.53 & 1063.14 \\
6282d124-7f24-11d8-8658-8df7c9751edc & 41 & 29.26 & 1053.58 \\
\vdots &  \vdots & \vdots & \vdots
    \end{tabular}
    \caption{Average scoring values.}
    \label{tab:average}
\end{table}

While used in practice to declare the winner in 
tournaments, for a ranking list does not reflect the
actual playing strength of the players, as there are many factors
that may have helped a lot. The number of games being played, 
the choice of opponents, the cards been given, etc. 

\subsection{Including Opponent Strength}

For a better assessment, an ELO rating 
system is needed, which combines 
playing results, strength of the opponents and luck in card deals. We propose a new one as, unfortunately, a widely accepted rating system 
does not exists for Skat, nor are other means of measuring 
playing strength used on selected servers accessible.
By the influence of chance computing ELO is more challenging 
than for Chess and similar deterministic games: 
there is no doubt, that even much weaker players can beat
much stronger ones rather easily when given 
a very good hand of cards~\cite{Duersch}.

It naturally plays a role how an ELO value is in relation to that of the opponent, the ELO quotient $EQ$, and how an own series result is in relation to the opponent’s one, the game quotient $GQ$. The ratio is $Y = GQ / EQ$.
 
Therefore, the first idea was to multiply the play value $S$ with factor $Y$ and use this result to update the ELO value. This lead to excessively large results,  
To dampen this effect we tried factor $\sqrt{Y}$. To avoid results far back in history tying down the current ELO value $R$, factor $f$ relates the weight of the earlier series $R \leftarrow (R \cdot f + S \sqrt{Y}) / (f+1)$, so that the old ELO value is $f$ times more important than the modified value $S \cdot \sqrt {Y}$ from the current series. Results with this ELO value, however, 
were not convincing.

Other ELO systems initiate the procedure with every player entering the race using a freely chosen starting rating, so one can base individual evaluation of a player to consider the ELO development of previous opponents. 

We think that keeping the data of all previous series individually for each player is not necessary. Due to the calculation, the influence of the "old" results fades continuously anyway and one, therefore, our system does not have to reevaluate the series that were played long ago.

We propose to compute the sum of Seeger-Fabian evaluation of played series $S=S_0+S_1+S_2$ and of the rating $R=R_0+R_1+R_2$.
Next, we calculate the expected values:
\begin{eqnarray*}
E_0 &=& R_0 \cdot S/R, \\
E_1 &=& R_1 \cdot S/R, \mbox{and} \\
E_2 &=& R_2 \cdot S/R.
\end{eqnarray*}
This has the effect that the sum of 
expected value is equal to the series and 
expected values are in the same relation as the rating of
the players.  This way outliers are avoided.
The update then is
\begin{eqnarray*}
R_0 & \leftarrow & R_0 + K\cdot (S_0-E_0), \\
R_1 & \leftarrow & R_1 + K \cdot (S_1-E_1), \mbox{and} \\
R_2 & \leftarrow & R_2 + K \cdot (S_2-E_2). 
\end{eqnarray*}
As an example consider the rating values $R_0=1500$, $R_1=750$, $R_2=750$ 
(assuming a ELO rating sum of $3000$ and a start value of $1000$), and a series with an 
actual scoring of $1200$, $800$, $800$.
Then $E_0=1400$, $E_1=700$, and $E_2=700$, The update yields new ranking
values  
$1500-K\cdot 200$,  $750+K\cdot 100$, and  $750+K \cdot 100$.
If we take $K=0.02$, then the new
rankings are $R_0=1496$, $R_1=752$, and $R_2=752$. We see that it is not necessarily advantageous for the better player to play against weaker ones. In contrast to his/her
clear victory he actually loses rating points, as s/he under-performs to the expectation.
In our initial experiments shown in Table~\ref{tab:elo} value $K=0.02$ leads to good results.
\begin{table}[t]
    \centering \scriptsize
    \begin{tabular}{ccc|ccc|ccc}
    $S_0$ & $S_1$ & $S_2$ & $E_0$ & $E_1$ & $E_2$ & $R_0$ & $R_1$ & $R_2$  \\ \hline
1387 & 1018 & 152 & 852.33 & 852.33 & 852.33 & 810.69 & 803.31 & 785.99  \\
501 & 1359 & 934 & 943.78 & 935.19 & 915.03 & 801.84 & 811.79 & 786.37  \\
637&1284&1125&1017.67&1030.30&998.04&794.22&816.86&788.91 \\
800&812&913&835.59&859.41&830&793.51&815.92&790.57 \\
1213&1221&965&1123.81&1155.54&1119.65&795.30&817.22&787.48 \\
1124&1445&994&1180.68&1213.24&1169.08&794.16&821.86&783.98 \\ 
1381&1187&938&1160.14&1200.60&1145.26&798.58&821.59&779.83 \\
693&374&1536&866.13&891.08&845.79&795.12&811.25&793.64 \\
460&1167&721&777.89&793.67&776.44&788.76&818.71&792.53 \\
871&15&1417&756.88&785.62&760.50&791.04&803.30&805.66 \\
\vdots &  \vdots & \vdots & \vdots &  \vdots & \vdots & \vdots &  \vdots & \vdots 
    \end{tabular}
    \caption{Evolution of ELO values with start value $800$.}
    \label{tab:elo}
\end{table}
We see that  $K$ has a significant 
influence not only on the spread of values but
also on the final standing of the players.

\subsection{Reducing Effect of Chance in the ELO System}

When talking to Skat experts we identified two main factors that are frequently
addressed in assessing the luck of players
in card games: strong hands and high-valued games.

\paragraph{Hand Strength}

For integrating chance into this ELO system one extreme is to avoid evaluating games with 100\% winning probability.
Instead, we normalize the strength of the current hand per game category with the mean hand strength (using win\-ning probabilities, the Kinback, or the more involved von-Stegen system). E.g,, if the mean is 8.72 and the current hand
is 9.5 we have a correction term of $c=1.09$, 
so that a score of 86 points is downgraded to 79.

If the values $c$ fall outside the interval $[0.5,2]$ we have clipped it to avoid extreme
behaviors by 
setting c = $\max \{ \min \{ c, 2 \}, 0.5\}.$
Algorithmically, we have

\begin{scriptsize}
\begin{lstlisting} 
    p = winningprob(hand0,skat1,skat2)
    np = normalprob(game);
    q = p/np; 
    q = (q < 0.5) ? 0.5 : (q > 2) ? q = 2 : q;
\end{lstlisting}
\end{scriptsize}

with \emph{winningprob} as the estimated probability of winning,
and the empirically derived partitioning of
games won

\begin{scriptsize}
\begin{lstlisting} 
normalprob(game) 
  if (game < 13) return 80.4;
  else if (game == 24) return 93.4;
  else if (game == 23) return 62.0;
  else if (game == 35) return 71.1;
  else if (game == 46) return 90.0;
  else if (game == 59) return 94.5;
\end{lstlisting}
\end{scriptsize}

Now every game is re-weighted.
Assuming 875 ELO for the declarer and 950 for the opponents, we derive another normalization factor of
1.086. Taken together, so that a score of 86 will be downgraded to 73. 

\paragraph{High-value Games}

There is another objective often attributed. In some series people score high with playing
only a few, but very high-valued games (usually Grands), for the others they do not bit. To avoid 
such over-exaggeration, one could adapt the score of the games. 
To bypass this incluence of chance we calculated the mean 
point score $\bar{x}$ of a game with $41.0$ (see Table~\ref{tab:avergamevalue}).
The implementation is simple.
\begin{scriptsize}
\begin{lstlisting} 
      if (CHANCEFACTOR2) gamevalue = 41;
      if (CHANCEFACTOR1) gamevalue = gamevalue / q;
\end{lstlisting}
\end{scriptsize}

\begin{table}[t]
    \centering \scriptsize
    \begin{tabular}{c|c|c|c}
Pos & \#Games & Sum of Values & Mean \\ \hline 
1 & 
29930 &
1264945 &
42.3 \\
2 &
26708 &
1069177 &
40.0 \\
3 &
27206 &
1101191 &
40.5 \\ \hline
1--3 &
83844 &
3435313 &
{\bf 41.0} \\ \hline  
    \end{tabular}
    \caption{Computing average game value.}
    \label{tab:avergamevalue}
\end{table}

Instead of the actual (or normalized) game value 
influencing the ELO score, for this chance breaking
we take the (normalized) game value $\bar{x}$ (or normalized,
$\bar{x}/q$). This might be seen as an extreme setting, as
giving lower- and higher-valued games the same influence in
the ELO formula, does not reflect the risk taken to achieve
the higher score. Of course many combinations of two 
adaptations to reduce the element of chance in the game are possible. 
In both cases, we use the adapted score
$s'_i = c_i \cdot s_i$, for each 
player $i=0,1,2$, and sum the 
values for a series score 
values $S'_i$. The $S_i$'s are then
into the above ELO update formula. We have
\begin{eqnarray*}
R_0 & \leftarrow & R_0 + K\cdot (S'_0-E_0), \\
R_1 & \leftarrow & R_1 + K \cdot (S'_1-E_1), \mbox{and} \\
R_2 & \leftarrow & R_2 + K \cdot (S'_2-E_2). 
\end{eqnarray*}

\subsection{Including Start ELO}
 
Finding an appropriate 
initial ELO ranking value. Voices differ
between lower value of 800 and a higher value of 1000 as a valid choice. 

A possible alternative
is to use the average Seeger score of some $n$ first played tables. In other words, for player $i$ use the initialization $R_i^{(n)} = \sum_{t=0}^n S^{(t)} / n$. Where the exponent shows the index of the table. 
To 
initialize the ELO ranking 
further games could also be taken into account, possibly after some 
rescaling to the Seeger score.

\subsection{Newcomer Progression}

For the ELO numbers in Chess it is well-established fact that younger, or 
newer players get a larger $K$, value
to be able to grow in their experience and
climb up the ranking more quickly. With a larger number of games and ELO value, the factor then decreases (from say $K=40$ to $K=10$).
This also compensates for an wrong assumption
on the playing strength in the initial ELO.
 
\section{Experiments}

The ELO ranking routine is embedded in our Skat AI~\cite{Edelkamp1,Edelkamp2,Edelkamp3} 
written in C++, compiled with
\verb\gcc\ version 4.9.2 (optimization level \verb\-O2\). As the formulas are straight-forward, we strongly believe that  a Skat web application with a world-wide ranking
list similar to \url{2700chess.com} can be derived easily. For the time being our 
linux executable program runs on 1 core of an
Intel Xeon Gold 6140 CPU @ 2.30GHz.

\subsection{ELO Ranking of Human Players}

We implemented the proposed ELO System for the game of Skat. As the input we selected a database of $83844$ expert games, played on a server in $2329$ series (tables) of $36$ games each. The data 
we process has anonymous (alias hashed, blind) player names and table indices, to avoid 
options for giving advantage to known Skat players.

For this series of human games, we derived
a proper ELO ranking, according to the above formulas and different parameterization for $K$,
the hand strength function or the inclusion of reducing the effect of card luck. On our computer this analysis takes about $4s$ mainly due to computing the winning probabilities. 
The outcome is a ELO highscore table.
Tables~\ref{tab:elorank1} 
and~\ref{tab:elorank2} 
show the derived list. 

\begin{table}[t]
    \centering \scriptsize
    \begin{tabular}{c|c|c|c}
No & P-No & Player Name (hashed) & Elo \\ \hline     
1 & P-2 & 550d2996ded442172116845433d09570 & 856.77 \\
2 & P-25 & bb01edd8abe48241a43b81af3895c925 & 854.80 \\
3 & P-12 &  542e18999d44ec8427481bf9b75c4bc6 & 853.42 \\
4 & P-27 & 693730a9db9143b7fc0a37d948b6fb76& 853.32 \\
5 & P-31 & e243ef566bac2607033c320e56ca3c & 853.06 \\
\vdots &  \vdots & \vdots 
    \end{tabular}
    \caption{ELO ranking for Skat on sample with $K=0.02$. }
    \label{tab:elorank1}
\end{table}

\begin{table}[t]
    \centering \scriptsize
    \begin{tabular}{c|c|c|c}
No & P-No & Player Name (hashed) & Elo \\ \hline     
1 & P-368 & eea6ea6035503ac4493cf46f12ced5b0 &        1063.61 \\
2 & P-616  & 7c36b944da8dcab82ed1e8b1822438cf &       1058.20 \\
3 & P-663 &  6d71ec67d159864149195c79f2cd289f  &      1055.64 \\
4 & P-60 &   fcfcbaa041c41d65ea2f0c64070f3af5   &     1017.28 \\
5 & P-26&    aa6ee7912aca31ea9d26371b539f3b43    &    1002.74 \\
\vdots &  \vdots & \vdots 
   \end{tabular}
    \caption{ELO ranking for Skat on sample with $K=0.05$. }
    \label{tab:elorank2}
\end{table}

\subsection{ELO Ranking of Skat AIs}

We have developed a strong AI for the Skat game~\cite{Edelkamp1,Edelkamp2,Edelkamp3}, capable to beat even advanced club players. So far, we have recorded a few mixed Human-AI games.

We could replay all the games with our Skat AI.
To accelerate the evaluation on multiple cores, 
we partitioned the input file into 
$9 \times 8388$ games and $1 \times 8352$,
and started the driver for the AIs 10 times.
The results are shown in Fig~\ref{tab:aresults}.
Looking at the numbers on can extract that
the AI wins $4132+4768$ games that the Humans
do not, while the Humans win $5273+1085$ that
the AIs do not. This hints that the AIs 
showing a better playing performance,  
reflected also in the Seeger score.
Table~\ref{tab:playperformance} partitions the
playing results along different game types, and
shows that the AI scores more wins in all of them.
The running time for the entire experiment 
was around 8h but would have been 78h, if we 
were not to partition the input file. For each game
this means about 3s for the AIs to finish.
Unfortunately, AI self-play of the bots 
does hardly help to illustrate an ELO evolution.
If there are three bots, which are all the same,
the increase and decrease is essentially the same,
given that the sum of the ELO values stays constant.
To show progress AIs of different strength have 
to be playing against each other.

 \begin{table}[t]
    \centering
\scriptsize    \begin{tabular}{c|ccc|r}
        & Human & Glassbox & AI &  \# Games \\ 
                & Wins & Wins & Wins &   \\ \hline
          &  0 & 0 & 0 &  4080   \\
           &  0 & 0 & 1 & 4132   \\
           &  0 & 1 & 0 &  297   \\
           &  0 & 1 & 1 & 4768  \\ 
           &  1 & 0 & 0 & 5273   \\
           &  1 & 0 & 1 & 9442  \\
           &  1 & 1 & 0 & 1085  \\
           &  1 & 1 & 1 & 52820   \\ \hline  
         Total Score &   &  &  & 995.67    \\  \hline    
    \end{tabular}
    \caption{Playing over 83T mixed trump games with AIs.
    } 
    \label{tab:aresults}
\end{table}

\begin{table}[t]
    \centering \scriptsize
    \begin{tabular}{c|cc|c}
&WonHuman & WonAI & \%WonKI \\ \hline
Suit &
38975 &
41041 &
105.3\%
\\
Grand &
26677 &
26984 &
101.2\% \\
Null &
1531 &
1586 &
103.6\% \\
NO &
1437 &
1551 &
107,9\% \\ \hline
Total &
68620 &
71162 &
103.7\% \\    
    \end{tabular}
    \caption{Playing performance AI in the 83T Skat games.}
    \label{tab:playperformance}
\end{table}

\subsection{Sensitivity Analysis} 
 
To study the change in the ELO distribution we varied the parameters 
for the elements of change $i_1$ (winning probability) $i_2$ and the $K$-factor. 
 
\begin{table}[t]
    \scriptsize
    \hspace{-0.3cm}\begin{tabular}{cc|c|ccc|cccc}
$i_1$ & $i_2$ & K & Max & Mean & Min & 10 & 25 & 75 & 90 \\ \hline
0 & 0 & .01 & 924.84 & 849.91 & 790.21 & 862.18 & 854.33 & 845.07 & 839.22\\
0 & 0 & .02 & 976.54 & 849.90 & 739.95 & 874.07 & 858.67 & 840.34 & 829.28\\
0 & 0 & .04 & 1038.27 & 850.33 & 652.66 & 893.69 & 866.58 & 831.44 & 810.01\\
0 & 0 & .08 & 1149.71 & 850.81 & 532.15 & 922.17 & 881.78 & 815.78 & 778.33\\
\hline
0 & 1 & .01 & 920.87 & 849.87 & 793.42 & 860.88 & 854.25 & 845.61 & 840.24\\
0 & 1 & .02 & 962.12 & 849.75 & 745.01 & 870.80 & 858.32 & 841.32 & 831.08\\
0 & 1 & .04 & 1018.33 & 849.49 & 664.03 & 889.39 & 866.21 & 833.93 & 814.13\\
0 & 1 & .08 & 1120.17 & 849.02 & 542.22 & 920.97 & 881.58 & 819.69 & 785.47\\
\hline
1 & 0 & .01 & 927.56 & 849.85 & 776.35 & 863.45 & 854.95 & 844.96 & 838.80\\
1 & 0 & .02 & 981.48 & 849.91 & 714.23 & 875.16 & 859.43 & 840.14 & 828.00\\
1 & 0 & .04 & 1045.39 & 850.27 & 617.78 & 893.19 & 867.97 & 830.71 & 808.44\\
1 & 0 & .08 & 1165.40 & 850.29 & 502.82 & 929.80 & 884.76 & 813.70 & 773.46\\
\hline
1 & 1 & .01 & 922.10 & 849.81 & 777.90 & 862.79 & 854.83 & 845.28 & 839.55\\
1 & 1 & .02 & 973.63 & 849.67 & 715.30 & 874.93 & 859.47 & 840.70 & 829.59\\
1 & 1 & .04 & 1036.14 & 849.30 & 614.48 & 895.34 & 868.56 & 832.52 & 809.81\\
1 & 1 & .08 & 1144.11 & 849.79 & 487.55 & 931.75 & 884.90 & 817.03 & 778.09\\
\hline
    \end{tabular}
    \caption{ELO distribution for chance and 
    change parameters. }
    \label{tab:elodist}
\end{table}

Table~\ref{tab:elodist} shows that 
the distribution of ELO values
for a growing $K$ i´widens so that the volatility factor
$K$ can be adapted to the needs.
For a world-wide ELO list, or for 
frequently played online platforms,
a smaller value of $K$ might be appropriate
for a moderate increase and decrease in value. For a tournament or club
championship, a larger value of $K$ will 
be better.
We see the effect for reducing the element of chance in the ELO numbers. 
We are still trying to 
measure the effect statistically. 


\subsection{Visualization of Individual ELOs}

\begin{figure}
    \centering
    \includegraphics[width=4.3cm]{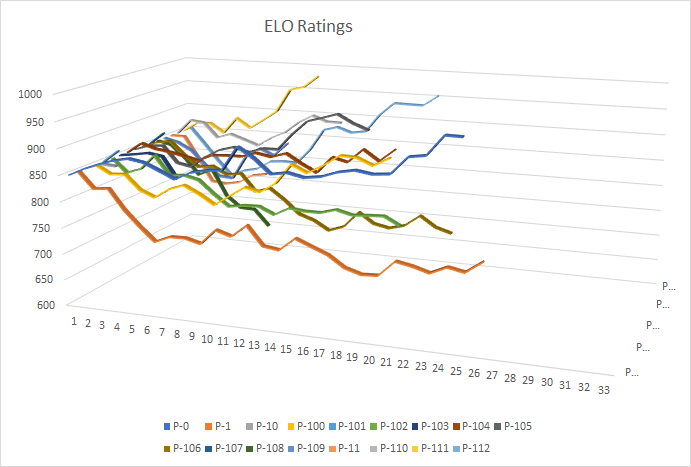}
    \includegraphics[width=4.3cm]{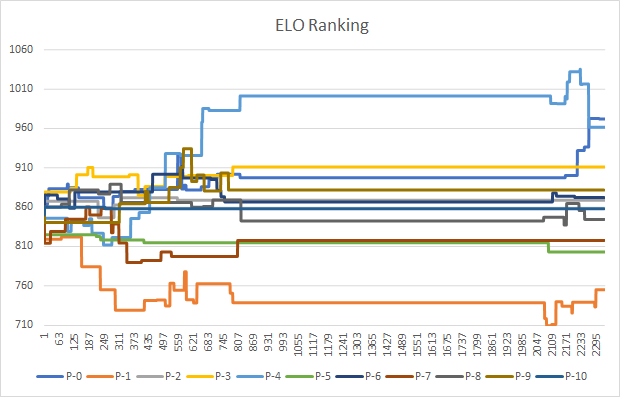}
    \caption{Players' performance in ELO chart: time-contracted view (left),
    time-expanded view (right). }
    \label{fig:elo1}
\end{figure}
 
 
Fig.~\ref{fig:elo1} 
illustrates the 
dynamic change in the ELO scoring values for some selected players. 
For the time-contracted view,
only the changes of ELO-values are stored, for the
time-expanded view, the table ID of
the game is taken to show 
the evolution of ELO of different players 
over time.
 
This visualization option could be provided
for the players (or tournament directors). 
Based on a mixed tournament and ELO rating,
fone could also award titles like National-,
International-, or Grandmasters, as done in Chess.

\begin{figure}[t]
    \centering
    \includegraphics[width=4.3cm]{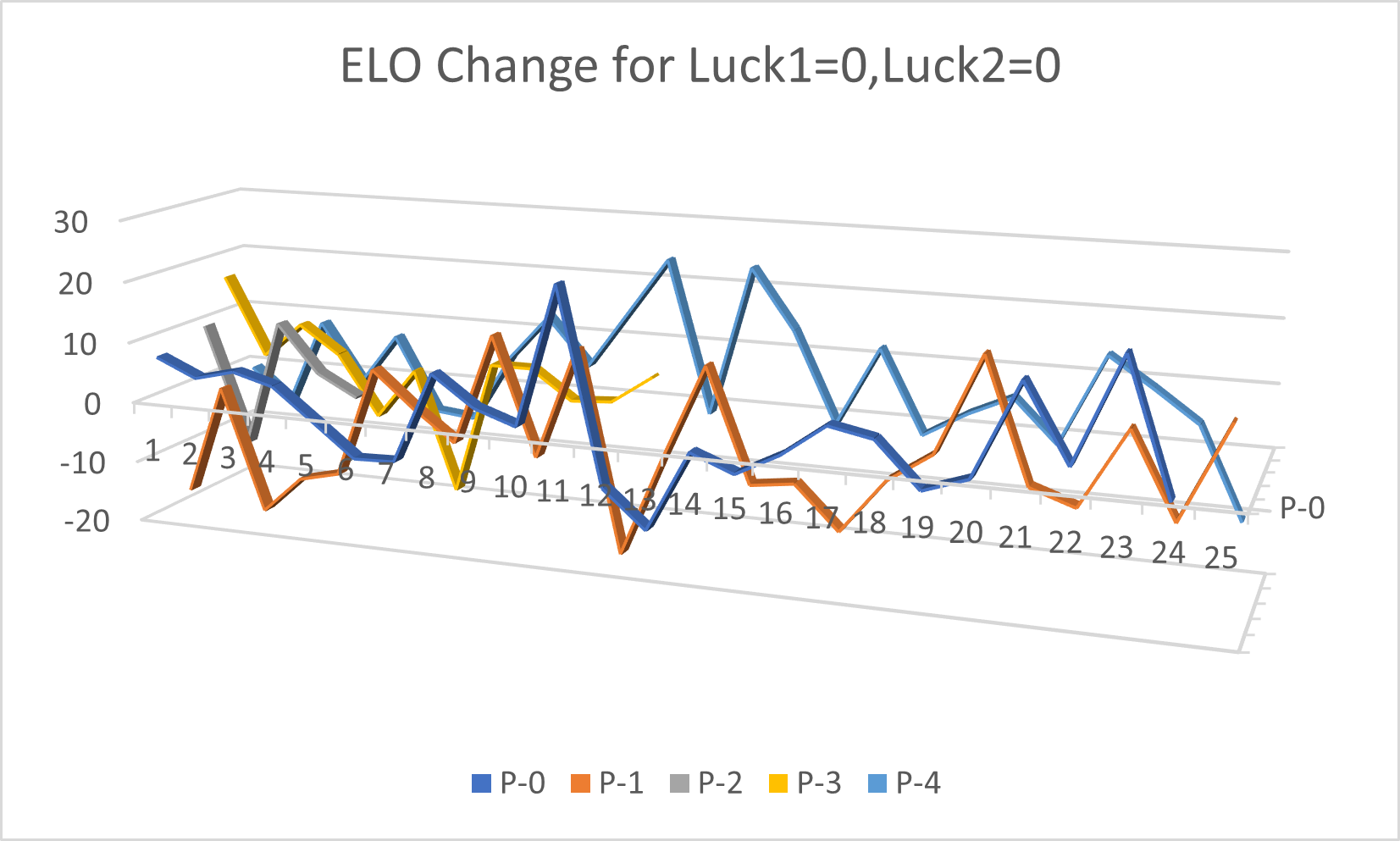}
    \includegraphics[width=4.3cm]{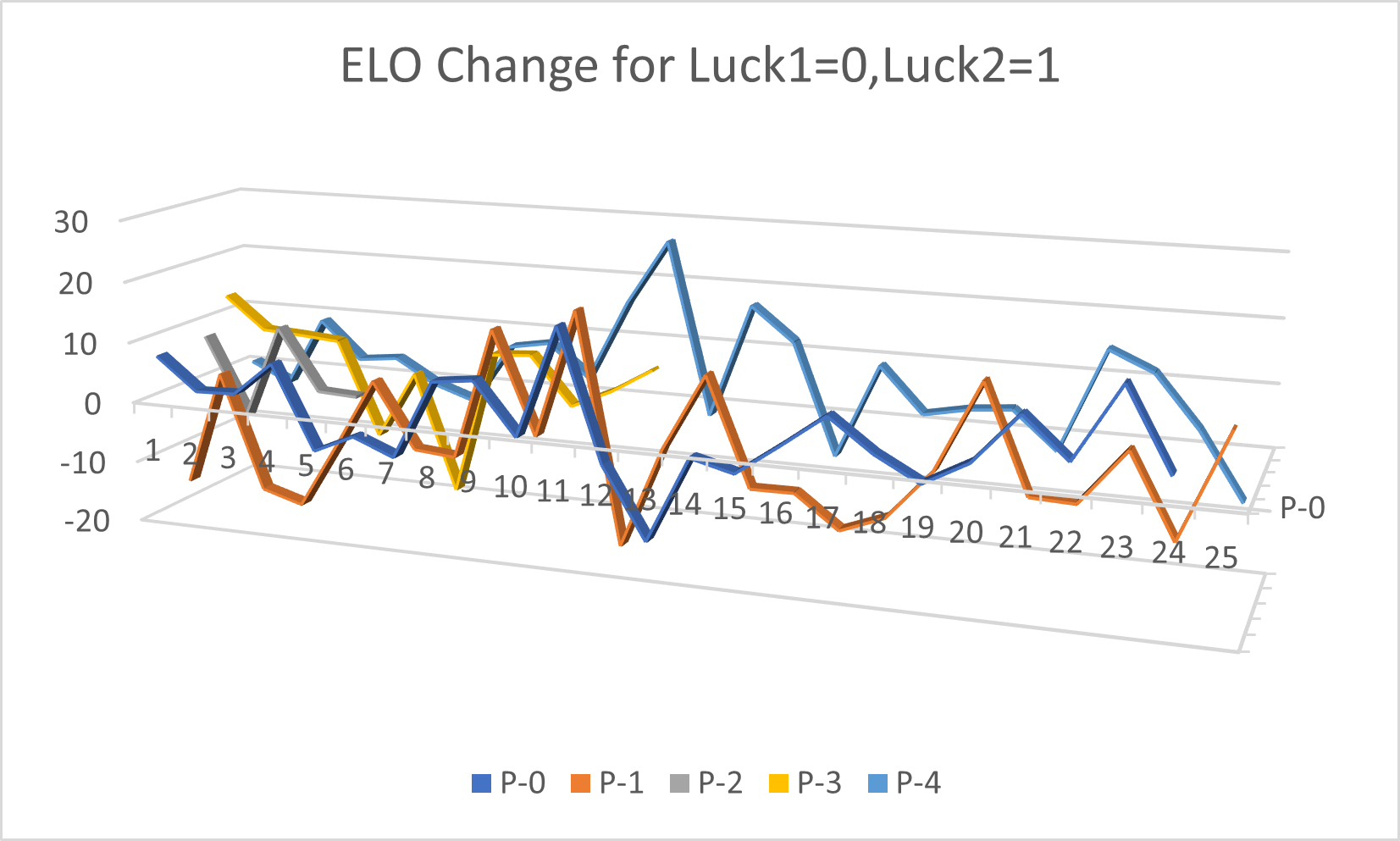}
    \includegraphics[width=4.3cm]{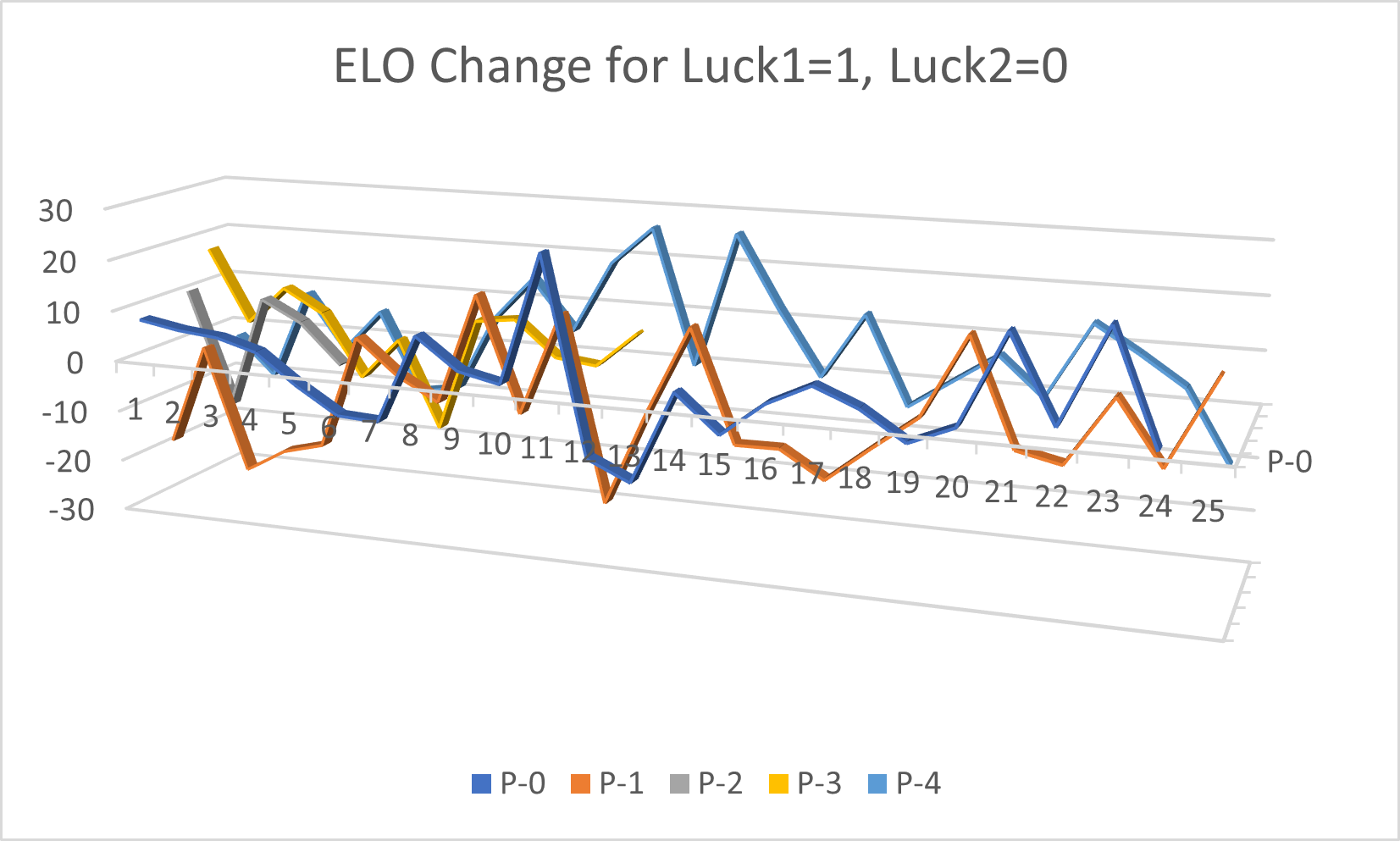}
    \includegraphics[width=4.3cm]{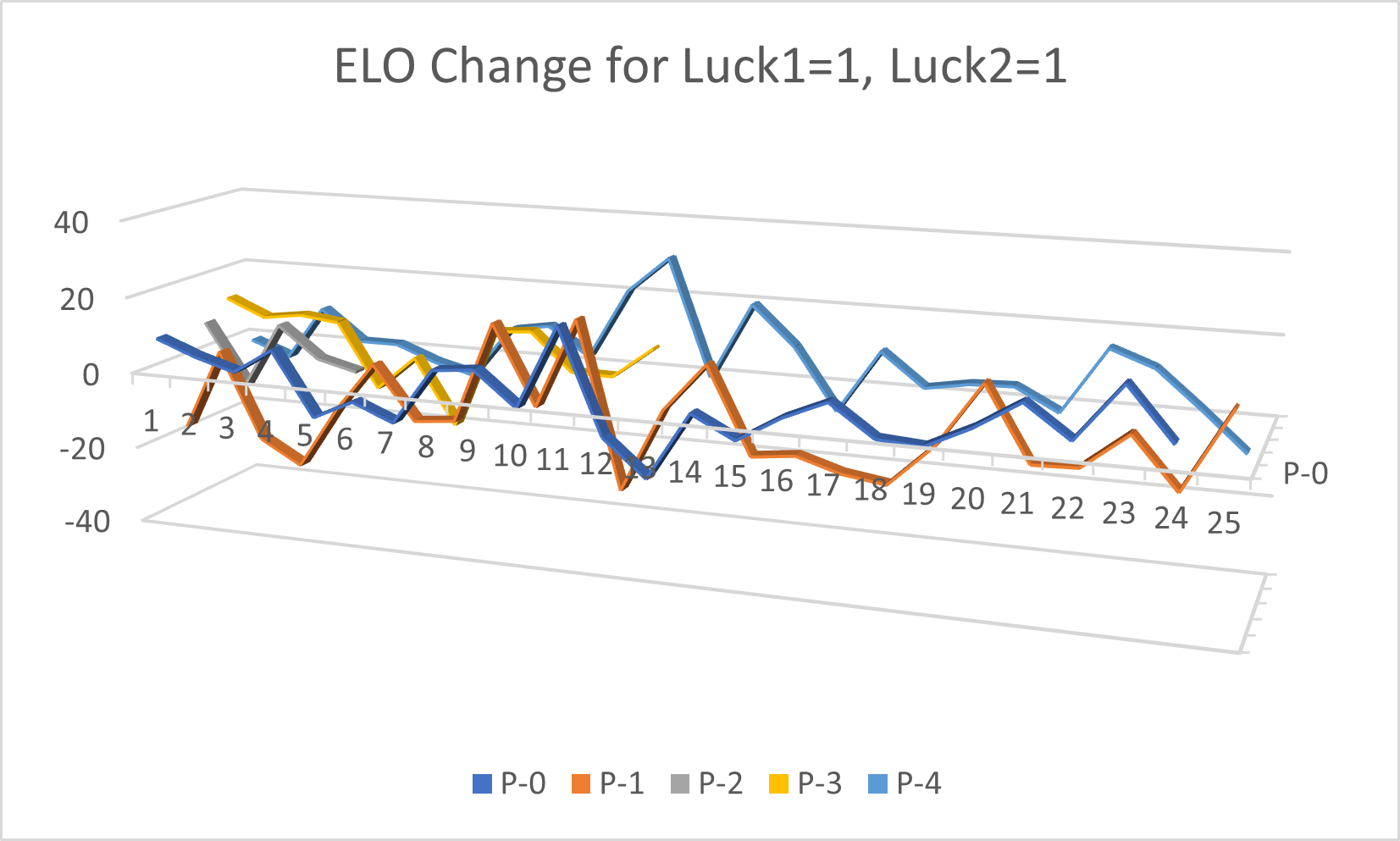}
    \caption{Sensitivity analysis for first five players, and different
    parameters for reducing effect of chance. }
    \label{fig:chance}
\end{figure}

In a time-contracted view,
Figure~\ref{fig:chance} shows the evolution of the change in the
ELO values. In the longer series we see that when including the
ELO formula adaptation for chance, the playing strength values show smaller
amplitudes, i.e., the curves become a bit smoother. This means that 
for this case the real strength of
a Skat player becomes visible faster, than without the inclusion
of the chance parameters.

\section{Conclusion}

We have introduced a novel ELO rating system for the game of Skat to rank human 
and automated players for their performance on a longer time-scale of play. 
The new ELO rating system includes both opponent strength and 
card luck, and aligns to the accepted Seeger  
scoring system used by the IPSA and DSKV in official 
tournaments. 

A transparent and fair ELO system for Skat and other games of chance is very much wanted 
to award players in tournaments or online play, and to set up live high-score and ranking
lists. We have started with a rather simple proposal for computing the ranking
that overcomes deficiencies of the
existing ELO system for two-player zero sum games that is based on predicting the
outcome based on the difference in ELO rating. Similar to the established ELO system, it is intuitive and can be parameterized. For Skat it offers a much fairer evaluation and ranking as simply than the usual approach of simply taking the average score. The system has
several advantages:
 
\begin{itemize} 
\item It is comparable to the extended Seeger scoring system used over decades by
DSKV and ISPA
\item We determined after how many series with some tolerance a 
\emph{fixpoint} of strength for a player is reached. 
\item Using the evaluation of reduced luck, while taking into account 
opponent strength, we reached a similar fixpoint after
a smaller number of series.
\item The ELO system itself is deterministic, the same set of games leads to the same ELO value.
\end{itemize}
 
Different to other researchers looking at ELO evolution to measure chance and skill in the game, the driving force of our research is to counter-balance the 
chance in order to
converge faster to the real playing strength. 
By the law of large numbers, one would expect that
the more games are played the less important the
influence of chance will be in the ELO rating. 
Besides the factors for chance, we saw this also depends 
on the volatility, namely factor $K$. 
 
Besides the Seeger scoring system as its base, there is not much in the proposal that is specific to Skat or to card game, so that we our ELO system can be used to evaluate competitions in AI for many other games of chance and partial information. One simply needs any existing and establish scoring system for a series of games. To reduce the factor for luck, a 
function estimating the strength of the initial position is needed for normalization. It might be learnt.

\medskip
 
\paragraph{Acknowledgement}
We thank Rainer G\"o{\ss}l for his 
contribution of expert-level Skat play driving the entire project, and 
Stefan Meinel for his deep mathematical insights. Both contributed to the
derivation of the proposed ELO formula.

\end{document}